\documentclass[times,twocolumn,final]{elsarticle}

\usepackage{style/prletters}
\usepackage{framed,multirow}
\usepackage{amsmath}
\usepackage{algorithm}
\usepackage{algorithmic}
\usepackage{subfigure}
\usepackage{booktabs}
\usepackage{url}
\usepackage{xspace}
\usepackage[pagebackref,breaklinks,colorlinks,citecolor=blue,linkcolor=blue,bookmarks=false]{hyperref}
\usepackage{balance}

\newcommand{\calD}{{\mathcal{D}}}
\newcommand{\calF}{{\mathcal{F}}}
\newcommand{\calL}{{\mathcal{L}}}
\newcommand{\calW}{{\mathcal{W}}}
\newcommand{\calX}{{\mathcal{X}}}

\newcommand{\norm}[1]{\| #1 \|}

\def\onedot{.\@\xspace}
\def\eg{\emph{e.g}\onedot}

\newcommand{\Eref}[1]{Eq.~(\ref{#1})}
\newcommand{\Fref}[1]{Fig.~\ref{#1}}
\newcommand{\Frefs}[1]{~\ref{#1}}  % Fref for short
\newcommand{\Tref}[1]{Table~\ref{#1}}
\newcommand{\Aref}[1]{Algorithm~\ref{#1}}

\journal{Pattern Recognition Letters}

\begin{document}

\ifpreprint
  \setcounter{page}{1}
\else
  \setcounter{page}{1}
\fi

\begin{frontmatter}

\title{Active Anomaly Detection based on Deep One-class Classification}

\author[1]{Minkyung \snm{Kim}}
\ead{mkkim1778@kaist.ac.kr}
\author[2]{Junsik \snm{Kim}}
\ead{jskim@seas.harvard.edu}
\author[3]{Jongmin \snm{Yu}\corref{cor1}}
\cortext[cor1]{Corresponding author.}
\ead{andrew.yu@kaist.ac.kr}
\author[1]{Jun Kyun \snm{Choi}}
\ead{jkchoi59@kaist.edu}

\address[1]{\small School of Electrical Engineering, Korea Advanced Institute of Science and Technology (KAIST), Republic of korea}
\address[2]{School of Engineering and Applied Sciences, Harvard University, USA}
\address[3]{Institute for IT Convergence, Korea Advanced Institute of Science and Technology (KAIST), Republic of korea}

\received{1 May 2013}
\finalform{10 May 2013}
\accepted{13 May 2013}
\availableonline{15 May 2013}
\communicated{S. Sarkar}

\begin{abstract}
Active learning has been utilized as an efficient tool in building anomaly detection models by leveraging expert feedback. 
In an active learning framework, a model queries samples to be labeled by experts and re-trains the model with the labeled data samples.
It unburdens in obtaining annotated datasets while improving anomaly detection performance.
However, most of the existing studies focus on helping experts identify as many abnormal data samples as possible, which is a sub-optimal approach for one-class classification-based deep anomaly detection.
In this paper, we tackle two essential problems of active learning for Deep SVDD: query strategy and semi-supervised learning method.
First, rather than solely identifying anomalies, our query strategy selects uncertain samples according to an adaptive boundary. 
Second, we apply noise contrastive estimation in training a one-class classification model to incorporate both labeled normal and abnormal data effectively.
We analyze that the proposed query strategy and semi-supervised loss individually improve an active learning process of anomaly detection and further improve when combined together on seven anomaly detection datasets.
\end{abstract}


\end{frontmatter}

%% ---------------------------------------------------------------------------------------------------------------------------------------
\section{Introduction}\label{sec:Introduction}
Anomaly detection (AD) techniques constitute a fundamental resource in various applications to identify observations that deviate considerably from what is considered normal.
In response to increasingly complex data at a large scale, deep learning-based AD has been actively researched and has shown high capabilities~\cite{pang2021deep}.
Since deep learning is based on a representation learned from a given dataset, most deep AD models aim to learn normality,
assuming that a dataset consisting of only normal samples is available.
Thereby, one-class classification (OCC)-based approaches are one of the representative approaches of deep AD~\cite{ruff2021unifying}.
However, in practice, it is expensive to prepare a training dataset consisting of normal samples as it requires per-sample inspection.
When unlabeled data is abundant, but manual labeling is expensive, active learning has been utilized as an efficient tool in building AD models while unburdening the labeling process and improving AD performance by leveraging feedback from domain experts
~\cite{barnabe2015active, das2016incorporating, lamba2019learning, lesouple2021incorporating, trittenbach2020overview, pimentel2020deep, zha2020meta}. 
There are two significant building blocks in active learning: a \textit{query strategy} for determining which data samples should be labeled and a \textit{semi-supervised learning method} for re-training AD models utilizing the labeled samples.

According to common AD scenarios that identify the top instances from a ranked list of anomalousness determined by an AD model,
most of the existing AD methods utilizing active learning select the top-1 or top-K instances as a query to be labeled
~\cite{barnabe2015active, das2016incorporating, das2017incorporating, lamba2019learning, lesouple2021incorporating}. 
However, this common but greedy query strategy is sub-optimal in training OCC-based AD, where samples with high uncertainty may be more informative than samples with high confidence. 
For an OCC-based AD model that learns an explicit decision boundary, data near the boundary is selected as a query to be labeled~\cite{gornitz2013toward, trittenbach2020overview}, i.e., uncertainty sampling. However, the OCC-based AD model usually relies on hyper-parameters to learn the boundary, and it is not easy to find the optimal value in the active learning framework since model re-training is repeated with the newly labeled data and remaining unlabeled data.

Another important part of active learning is incorporating labeled and unlabeled samples into training, \emph{i.e.}, semi-supervised learning.
The simple semi-supervised learning method for OCC-based AD is to exclude labeled abnormal samples while regarding both labeled normal and unlabeled samples as normal samples during training~\cite{barnabe2015active}.
However, completely excluding labeled abnormal samples in training is a loss of information.
Later works~\cite{ruff2020deep, lesouple2021incorporating} incorporate labeled abnormal samples into semi-supervised learning to learn discriminative characteristics between normal and abnormal samples. 
However, they still do not discriminate between labeled normal and unlabeled samples in the training phase with the assumption that most of the unlabeled data are also normal. This indiscriminate use of samples is another loss of information, as labeled normal samples should have less uncertainty than unlabeled samples.

In this work, we tackle two essential problems of active learning for OCC-based deep AD.
First, we propose an uncertainty-based query strategy without learning an explicit decision boundary. To this end,
we search uncertain regions in a feature space where the density of normal and abnormal samples is expected to be similar and query samples from the searched regions. 
In OCC-based approaches, the anomaly score is computed by the distance from one fixed coordinate called the center in a latent space.
Therefore, we search for a boundary where a similar number of normal and abnormal samples are located nearby.
The searched boundary, called the \textit{adaptive boundary}, is adjusted iteratively by the ratio of abnormal samples queried in the previous active learning stage.
In other words, instead of learning the explicit boundary with hyper-parameters, we utilize labeling information as feedback for the adaptive boundary.
For the semi-supervised learning method, unlike the previous work discriminating only labeled abnormal samples,
we apply noise contrastive estimation (NCE)~\cite{gutmann2010noise} on labeled normal samples to contrast them from labeled abnormal samples.
We empirically validate on seven anomaly detection datasets that the proposed query strategy with an adaptive boundary and the NCE-based semi-supervised learning method are effective when applied individually but even stronger when combined together.

%% ---------------------------------------------------------------------------------------------------------------------------------------
\section{Related work}\label{sec:Related_work}
\subsection{OCC-based anomaly detection}\label{sec:occ-ad}
One-class classification (OCC) identifies objects of a specific class amongst all objects by primarily learning from a training dataset containing only the objects of the target class~\cite{perera2021one}. 
It detects whether new instances conform to the training dataset or not. 
On the other hand, the objective of anomaly detection (AD) is to separate abnormal data from a given dataset, which is a mixture of normal and abnormal data without ground truth annotations. 
Despite this difference, OCC has been widely applied to AD to learn the normality of a given dataset. 

The most widely known shallow methods for OCC-based AD are One-Class SVM (OC-SVM)~\cite{scholkopf2001estimating}, and Support Vector Data Description (SVDD)~\cite{tax2004support}. 
They find a hyperplane and a hypersphere enclosing most of the training data, respectively.
They deal with abnormal samples through hyper-parameters that allow some points to be put outside the estimated region. 
On the other hand, the works~\cite{gornitz2013toward, tax2004support} incorporate labeled data into SVDD to refine a boundary in a semi-supervised way.
Tax \emph{et al.}~\cite{tax2004support} enforce labeled abnormal samples to lie outside the hypersphere. 
Gornitz \emph{et al.}~\cite{gornitz2013toward} add one more constraint for labeled normal samples to be inside the hypersphere (See \cite{trittenbach2020overview} for more details).

The shallow methods have been the solid foundation for OCC-based deep AD (OCC-DAD)~\cite{ruff2018deep, chalapathy2018anomaly, ruff2020deep, zhang2021anomaly}.
Most of them replace the shallow models with deep neural networks while using the same objective functions as in OC-SVM or SVDD.
However, they commonly assume all training data are normal. 
We consider them as semi-supervised learning methods that exploit only labeled normal data while excluding labeled abnormal data.
On the other hand, Ruff \emph{et al.}~\cite{ruff2020deep} propose a method to incorporate labeled anomalies into \textit{One-Class} Deep SVDD~\cite{ruff2018deep}, a prevailing method in OCC-DAD.
In addition, another line of OCC-DAD utilizes a generative adversarial network to exploit distorted anomalies adversarially~\cite{sabokrou2018adversarially}.
However, it is difficult to identify the generated samples and guarantee that they resemble true anomalies.

\subsection{Active learning for OCC-based AD}
According to the survey~\cite{trittenbach2020overview}, the existing query strategies (QS) for OCC-based AD can be classified into three categories. 
First, data-based QSs~\cite{ghasemi2011active} utilize data statistics such as posterior probabilities of samples; therefore, they require labeled observation even at the initial stage of active learning. 
The second category is model-based QSs~\cite{barnabe2015active, lesouple2021incorporating, gornitz2013toward}, to which the following two QSs belong. 
\textit{High-Confidence}~\cite{barnabe2015active, lesouple2021incorporating} selects samples that least match the normal class. 
It shares the view of common anomaly detection scenarios that identify the top instances from a ranked list of anomaly scores;
thereby, it is a common and widely used QS in active learning-based AD.
However, a recent study~\cite{zha2020meta} shows that this greedy choice could be sub-optimal since some low-ranked samples could be more informative.
The other QS in this category, \textit{Decision-Boundary}~\cite{gornitz2013toward}, selects samples closest to a decision boundary. 
However, OCC-based AD models rely on hyper-parameters to learn an explicit boundary.
The third category is hybrid QSs combining the former two categories. 
Since we tackle a dataset without any initial annotations in this work, we only consider the model-based QSs. 
Queried samples are utilized according to the semi-supervised learning methods described in Section~\ref{sec:occ-ad}.

%% ---------------------------------------------------------------------------------------------------------------------------------------
\begin{figure*}[t]
    \centering
    \includegraphics[width=0.9\textwidth]{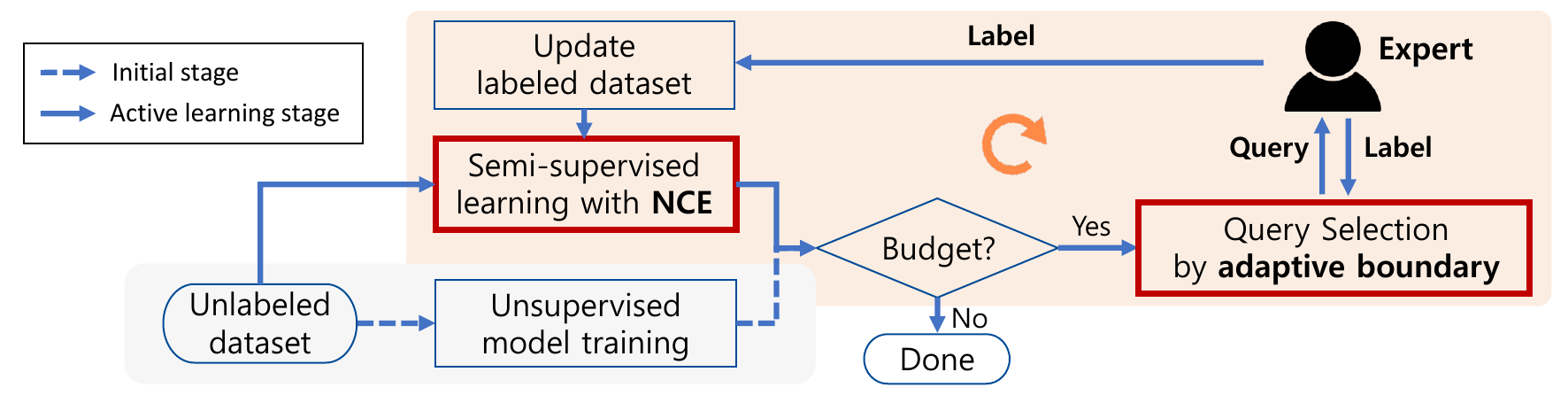} 
    \caption{Active learning process for deep one-class classification}
    \label{fig:framework}
\end{figure*}
\begin{figure*}[t]
\centering
\subfigure[Dataset: Ionosphere (35.9\%)\label{fig:observation_ionosphere}]
    {\includegraphics[width=0.33\textwidth]{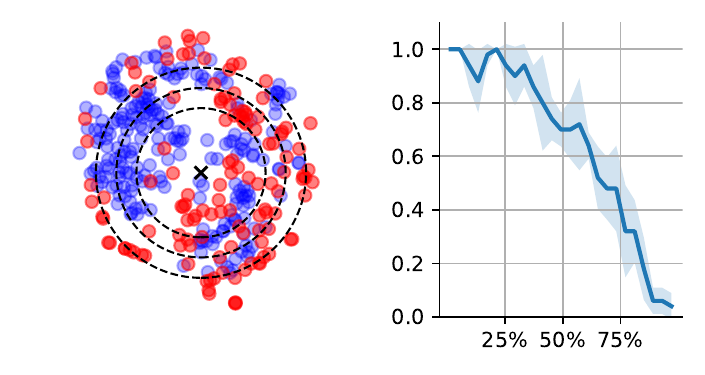}}
\subfigure[Dataset: Arrhythmia (14.6\%)\label{fig:observation_arrhythmia}]
    {\includegraphics[width=0.33\textwidth]{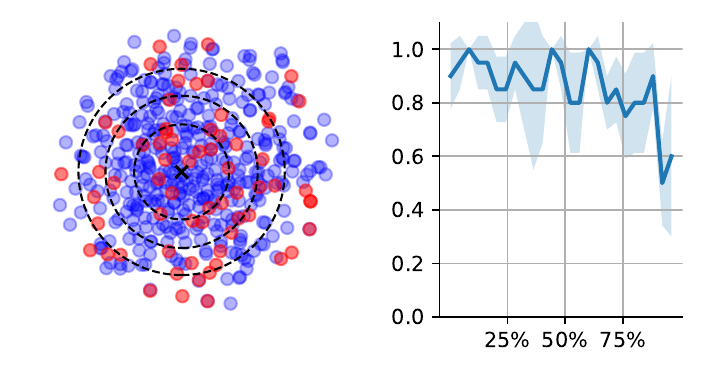}}
\subfigure[Dataset: Glass (4.2\%)\label{fig:observation_glass}]
    {\includegraphics[width=0.33\textwidth]{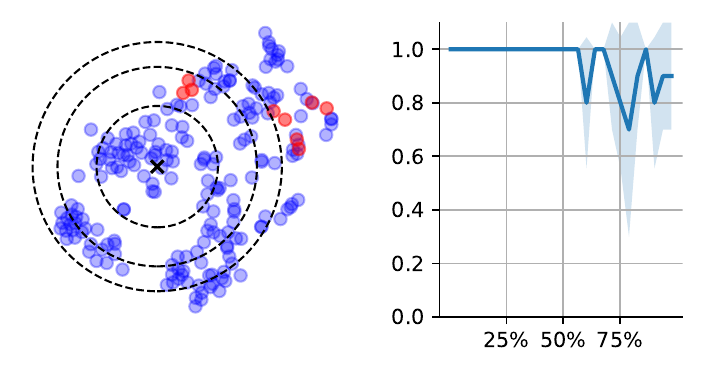}}
\caption{
    (Left) Examples of contaminated normality in a feature space of \textit{One-Class} Deep SVDD.
    The numbers in parentheses indicate anomaly ratios.
    We visualize the latent feature space mapped to two-dimensional space via t-SNE~\cite{van2008visualizing}.
    Blue (red) dots denote normal (abnormal) samples.
    The concentric circles represent the boundary containing 25\%, 50\%, and 75\% of the total data from the center of the hypersphere.
    (Right) The graph plots a ratio of normal samples among 6 samples closest to varying boundaries. 
    The X-axis represents a location of varying boundaries, and Y-axis represents a ratio of normal samples.
    The averaged ratios computed over 5 seeds are shown with a standard deviation represented in a shaded area.}
\label{fig:observations}
\end{figure*}

\section{Preliminary}\label{sec:Preliminaries}
We provide a brief introduction to Deep SVDD (DSVDD)~\cite{ruff2018deep}, which is used as a base model in this work.
Given a training dataset $\calD=\{\boldsymbol x_1, \cdots,\boldsymbol x_n\}$, where $\boldsymbol x_i \in \mathbb{R}^d$,
DSVDD maps data $\boldsymbol x\in\calX$ to a feature space $\calF$ through $\phi(\cdot;\calW):\calX \rightarrow \calF$ while gathering data around the center $\boldsymbol c$ of a hypersphere in the feature space, where $\calW$ denotes the set of weights of the neural network $\phi$.
Then, the sample loss $l$ for DSVDD is defined as follows:
\begin{equation}
    \label{eqn:dsvdd-sb}
    l(\boldsymbol x_i) = \nu \cdot R^2 + \textrm{max}(0, \norm{\phi(\boldsymbol x_i; \calW) - \boldsymbol c}^2-R^2).
\end{equation}
For simplicity, we omit regularizers in the rest of this section.
The hyper-parameter $\nu$ controls the trade-off between the radius $R$ and the amounts of data outside the hypersphere.
In the case that most of the training data are normal, the above sample loss can be simplified, which is defined as follows:
\begin{equation}
    \label{eqn:dsvdd-oc}
    l(\boldsymbol x_i) = \norm{\phi(\boldsymbol x_i; \calW) - \boldsymbol c}^2.
\end{equation}
These two versions of DSVDD are called \textit{soft-boundary} DSVDD and \textit{One-Class} DSVDD.
An anomaly score in both models is measured by the distance between the center and a data point (\Eref{eqn:dsvdd-score}).
\begin{equation}
    \label{eqn:dsvdd-score}
    s(\boldsymbol x_i;\calW) = \norm{\phi(\boldsymbol x_i;\calW)-\boldsymbol c}^2.
\end{equation}

%% ---------------------------------------------------------------------------------------------------------------------------------------
\section{Method}\label{sec:Method}
Since we consider a dataset without any annotations, a base model, Deep SVDD (DSVDD), is trained with the average of sample losses (\Eref{eqn:dsvdd-sb} or \Eref{eqn:dsvdd-oc}) across all the training data at the initial stage of active learning.
In subsequent stages, querying, labeling, and semi-supervised learning are repeated until an annotation budget is exhausted (\Fref{fig:framework}).
We consider that oracle labels queried samples. 
We denote the number of queried samples in each stage as $B$, 
the number of active learning stages as $T$,
a set of queried normal and abnormal samples at stage $t$ as $Q_N^t$ and $Q_A^t$,
a set of total labeled normal and abnormal samples as $L_N$ and $L_A$ and a set of remaining unlabeled samples as $U$.

\subsection{Uncertainty sampling with adaptive boundary}
Our query strategy with an adaptive boundary (AB) selects data samples closest to an adaptively searched boundary.
It is similar to Decision-Boundary in terms of performing uncertainty sampling. 
AB is motivated by the contaminated normality of DSVDD. 
Although DSVDD learns normality in feature space as a compact hypersphere enclosing normal samples, abnormal samples mixed in a dataset easily contaminate the normality (the left side of each figure in \Fref{fig:observations}).
It happens in both versions of DSVDD.
However, despite the contamination, we empirically observe that relatively more normal samples are located near the sphere’s center (the right side of each figure in \Fref{fig:observations}).
This phenomenon can be explained by the Effective gradient~\cite{xia2015learning}, which measures how much the error is reduced for a single datum, \emph{i.e.}, the projection of the overall gradient $\bar{g}$ on the direction of a single datum’s gradient $g_i$:
\begin{equation*}
    \begin{split}
    g_i^{\textrm{effect}} &= \frac{<g_i, \bar{g}>}{|g_i|} = |\bar{g}|cos\theta(g_i, \bar{g}), \\
    \bar{g} &= \frac{d\,\frac{1}{n}\sum_{i=1}^{n}l(\boldsymbol x_i)}{d\phi}=\frac{1}{n}\sum_{i=1}^{n}g_i = \frac{2}{n}\sum_{i=1}^{n}(\phi(\boldsymbol x_i;\calW)-\boldsymbol c),
    \end{split}
\end{equation*}
where $\theta(g_i, \bar{g})$ is the angle between the two vectors.

\begin{figure}[t]
    \centering
    \includegraphics[width=0.9\columnwidth]{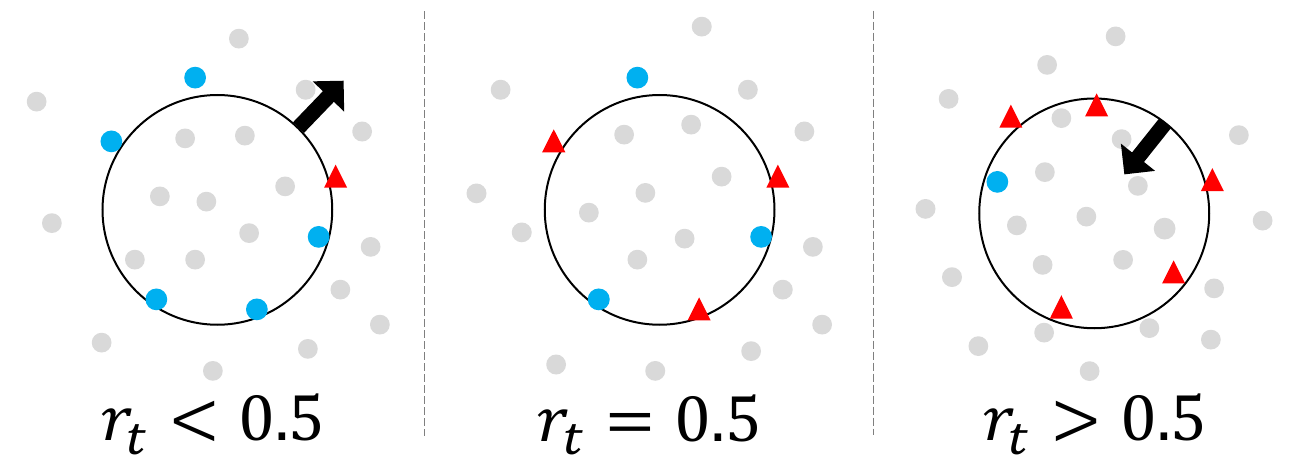} 
    \caption{
    Concept of adaptive boundary (AB) search.
    In each query stage, an AB is updated based on the ratio of queried abnormal samples $r_t$ at the previous stage.
    }
    \label{fig:rt}
\end{figure}

Following the observation, we assume that there exists a boundary (concentric spheres) in which the true label ratio of $B$ number of data closest to the boundary becomes half; and we consider those samples close to the boundary have high uncertainty of being judged as normal or abnormal by the model.
Our goal is to find the boundary where highly uncertain samples are nearby, which we call AB.
For searching AB, we use the ratio of abnormal samples among queried samples at the previous active learning stage. 
We denote the ratio as $r$.
The idea is that when the ratio $r$ is high (low), the AB moves in a direction towards (away from) the center (\Fref{fig:rt}). 
For the degree of movement, we utilize the difference between $r$ and its expected ratio of 0.5. 

We denote the ratio of data enclosed by AB as $q$. 
Suppose $q_t$ is 0.8 at active learning stage $t$. In that case, AB is the boundary of the hypersphere enclosing 80\% of the total data. 
We denote the ratio of queried abnormal samples in stage $t$ as $r_t$, which means $|Q_A^t|/B$.
The AB is updated as follows\footnote{$q_{t+1} = (q_{t} + q_{t-1})/2, \; \textrm{when} \; q_{t}=1$.}:
\begin{equation}
    \label{eqn:ab}
    q_{t+1} = q_t - \delta(q_t, r_t), 
\end{equation}
where $\delta(q_t, r_t)$ is the degree of movement derived from the proportional equation (\Eref{eqn:delta}).
We let $\delta(q_t, r_t)$ decrease as the AB widens, \emph{i.e.}, $q_t$ increases, by assuming that the ratio of abnormal samples from the current AB to the maximum boundary, \emph{i.e.}, $q_t=1$, increases linearly. We assume $r_t$ is 1 when $q_t$ is 1.
\begin{equation}
    \label{eqn:delta}
    (1-q_t) : (1-0.5) = \delta(q_t, r_t) : (r_t-0.5).
\end{equation}

\subsection{Joint training with NCE}
We propose to use noise contrastive estimation (NCE) as effective semi-supervised learning (SSL) method.
In utilizing NCE, we pay attention to a fundamental role of an anomaly score function, which assigns a lower score to normal data than abnormal data. 
We consider this role analogous to NCE, which estimates a model describing observed data while contrasting them with noise data.
Furthermore, an anomaly score function can be regarded as an unnormalized and parameterized statistical model. 
Therefore, we propose the NCE-based SSL method by setting the anomaly score of labeled normal and abnormal data as the estimation target and noise data.
Although noise data leveraged by the original NCE is artificially sampled from a known distribution, we assume labeled abnormal data are sampled from the true distribution of abnormal data while querying. 
The total loss $\calL$ for NCE-based SSL is as follows:
~
\begin{equation}
    \label{eqn:nce-ssll}
    \begin{split}
    \calL &= \frac{1}{|U\cup L_N \setminus Q_A^{PS}|}\sum_{\boldsymbol x_i \in U\cup L_N \setminus Q_A^{PS}} \norm{\phi(\boldsymbol x_i; \calW) - \boldsymbol c}^2 \\
    &- \frac{1}{|L_N \cup L_A|}\sum_{\boldsymbol x_i \in L_N}\sum_{\boldsymbol x_j \in L_A} \ln [1-h(\boldsymbol x_i, \boldsymbol x_j; \calW)],
    \end{split}
\end{equation}
\begin{equation}
    h(\boldsymbol x_i, \boldsymbol x_j; \calW) = \frac{s(\boldsymbol x_i; \calW)}{s(\boldsymbol x_i; \calW)+s(\boldsymbol x_j; \calW)},
\end{equation}
where $Q_A^{PS}$ denotes pseudo-abnormal samples. $Q_A^{PS}$ is defined by labeled abnormal samples as follows:
\begin{equation}
    \label{eqn:pseudo}
    Q_A^{PS} = \{\boldsymbol x_i | s_i > \text{median}(\{s_j| \boldsymbol x_j \in L_A\}), \\ \boldsymbol x_i \in U\}.
\end{equation}

For brevity, we denote $s(\boldsymbol x_i; \calW)$ as $s_i$.
The total loss is described using \textit{One-Class} DSVDD as a base model, but \textit{soft-boundary} DSVDD can also be applied by replacing the sample loss.
We outline the active learning process in \Aref{alg:algorithm}.

\begin{algorithm}[t]
\caption{Active anomaly detection based on Deep SVDD}
\label{alg:algorithm}
\textbf{Input}: Unlabeled dataset $\calD=\{\boldsymbol x_1, \cdots, \boldsymbol x_n\}$ \\
\textbf{Parameter}: B, T, $q_1$\\
\textbf{Output}: Anomaly scores
\begin{algorithmic}[1]
\STATE $U \leftarrow \calD$, $L_N \leftarrow \emptyset$, $L_A \leftarrow \emptyset$
\STATE Model training with $\frac{1}{n} \sum_{i}^{n} l(\boldsymbol x_i)$ (\Eref{eqn:dsvdd-sb} or \Eref{eqn:dsvdd-oc})
\FOR{t=1 to T}
\STATE Querying $B$ data samples closest to \\adaptive boundary $q_t$ ($B=|Q_N^t \cup Q_A^t|$)
\STATE $L_N \leftarrow L_N \cup Q_N^t$, \quad $L_A \leftarrow L_A \cup Q_A^t$
\STATE $U \leftarrow U \setminus (Q_N^t \cup Q_A^t)$
\STATE NCE-based semi-supervised learning (\Eref{eqn:nce-ssll})
\STATE Update adaptive boundary (\Eref{eqn:ab}) %based on $q_t$ and $r_t$ (\Eref{eqn:ab})
\ENDFOR
\end{algorithmic}
\end{algorithm}

%% ---------------------------------------------------------------------------------------------------------------------------------------
\section{Experiment}\label{sec:Experiment}
\subsection{Experiment settings}
\noindent\textbf{Datasets and evaluation metrics.}\
\emph{Anomaly detection benchmarks}~\cite{rayana2016odds} and \emph{NSL-KDD}~\cite{nsl-kdd2018} for network intrusion detection consist of multivariate tabular datasets. 
As shown in \Tref{tab:data}, each dataset has $n$ data samples with $d$ number of attributes.
The datasets used in this paper have a wide range of anomaly ratios in a dataset from a minimum of 2.9\% to a maximum of 46.5\%.
We use the area under the receiver operating characteristic curve (AUC) measured from anomaly scores as an evaluation metric.

\begin{table}[t]
    \caption{Anomaly detection benchmarks.}
    \label{tab:data}
    \centering
    \begin{tabular}{cccc}
    \toprule
    Dataset         & n             & d         & \# anomalies (\%) \\  \midrule
    NSL-KDD    & 125,973  & 41   & 46.5         \\
    Ionosphere      &     351       & 33        & 35.9              \\
    Arrhythmia      &     452       & 274       & 14.6              \\
    Cardio          &   1,831       & 21        &  9.6              \\
    Mnist           &   7,603       & 100       &  9.2              \\
    Glass           &     214       & 9         &  4.2              \\
    Optdigits       &   5,216       & 64        &  2.9              \\
    \bottomrule
    \end{tabular}
\end{table}

\begin{figure*}[t]
\centering
\subfigure[SSL method: OC\label{fig:agg-oc-a}]{\includegraphics[width=0.33\textwidth]{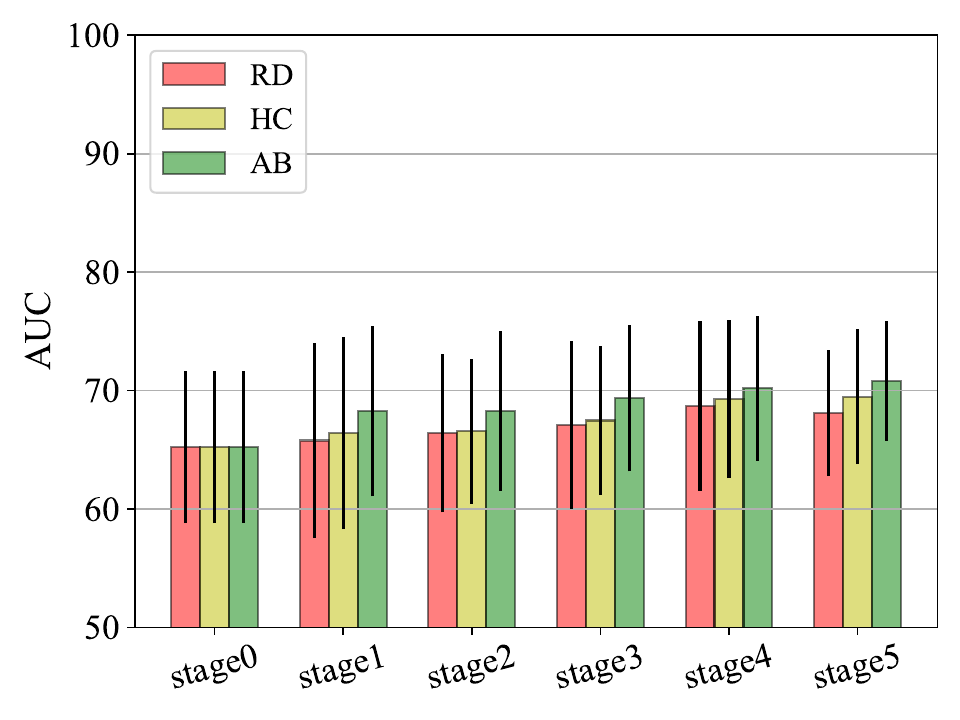}}
\subfigure[SSL method: DSAD\label{fig:agg-oc-b}]{\includegraphics[width=0.33\textwidth]{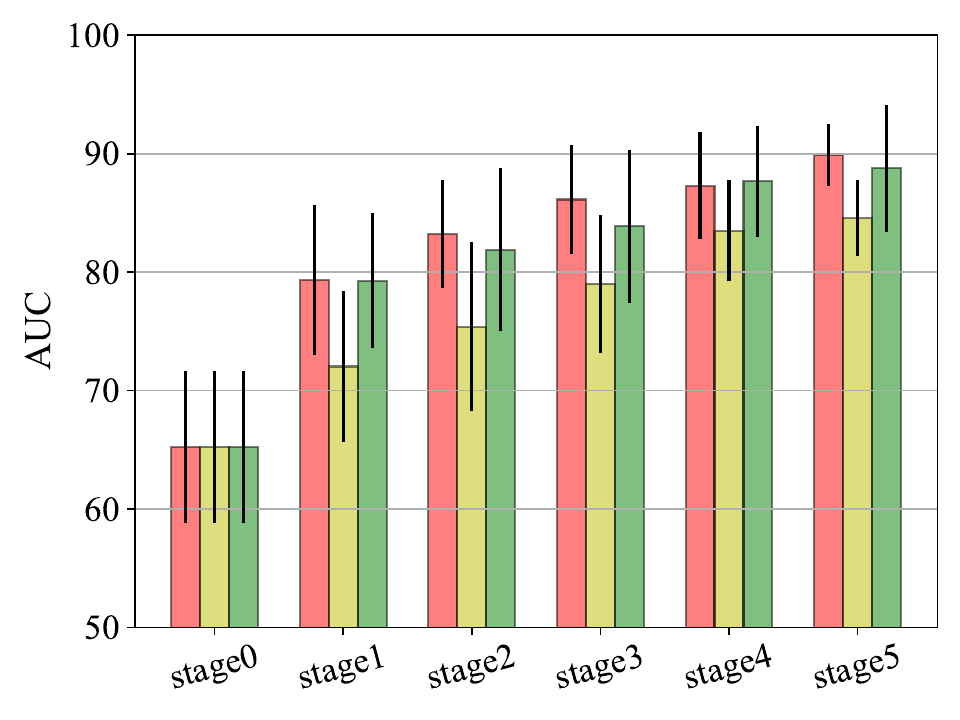}}
\subfigure[SSL method: OC-NCE\label{fig:agg-oc-c}]{\includegraphics[width=0.33\textwidth]{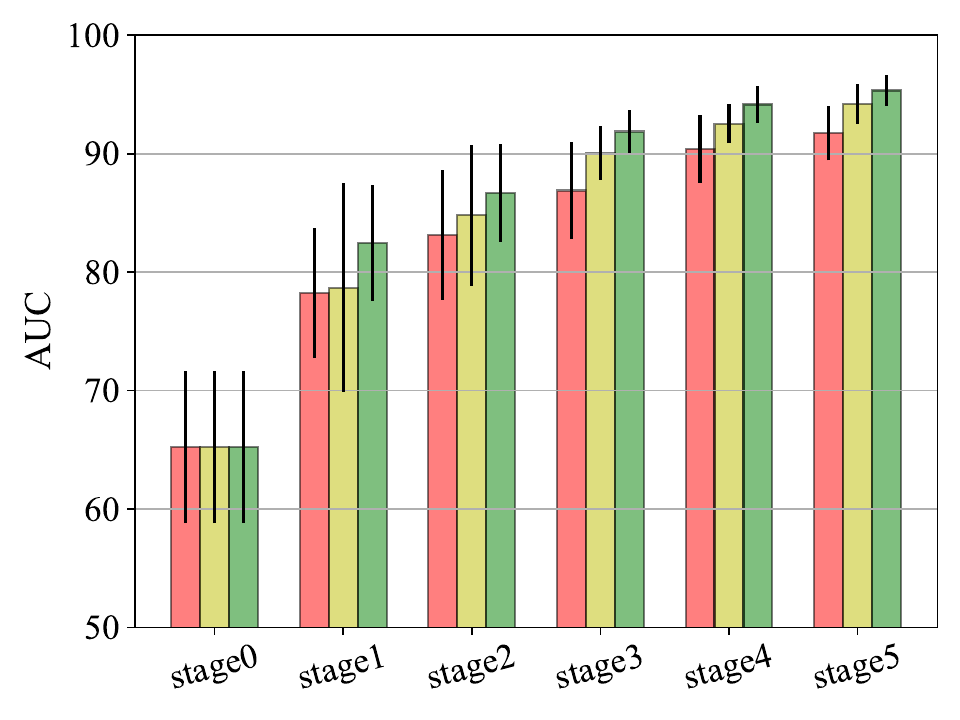}}
\linebreak
\subfigure[QS: Random\label{fig:agg-oc-d}]{\includegraphics[width=0.33\textwidth]{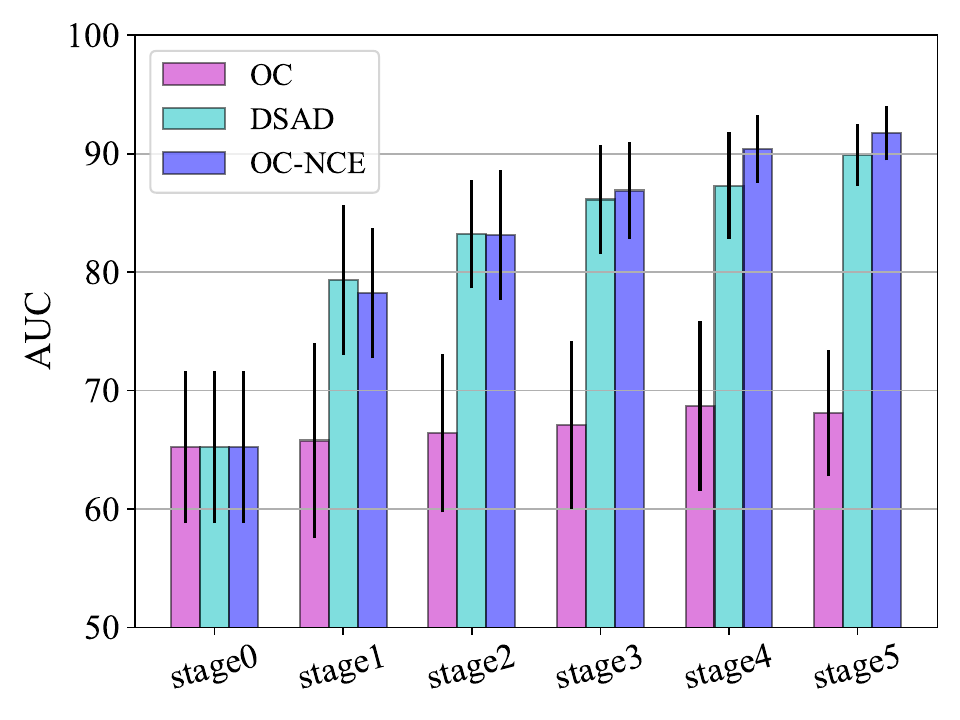}}
\subfigure[QS: High confidence\label{fig:agg-oc-e}]{\includegraphics[width=0.33\textwidth]{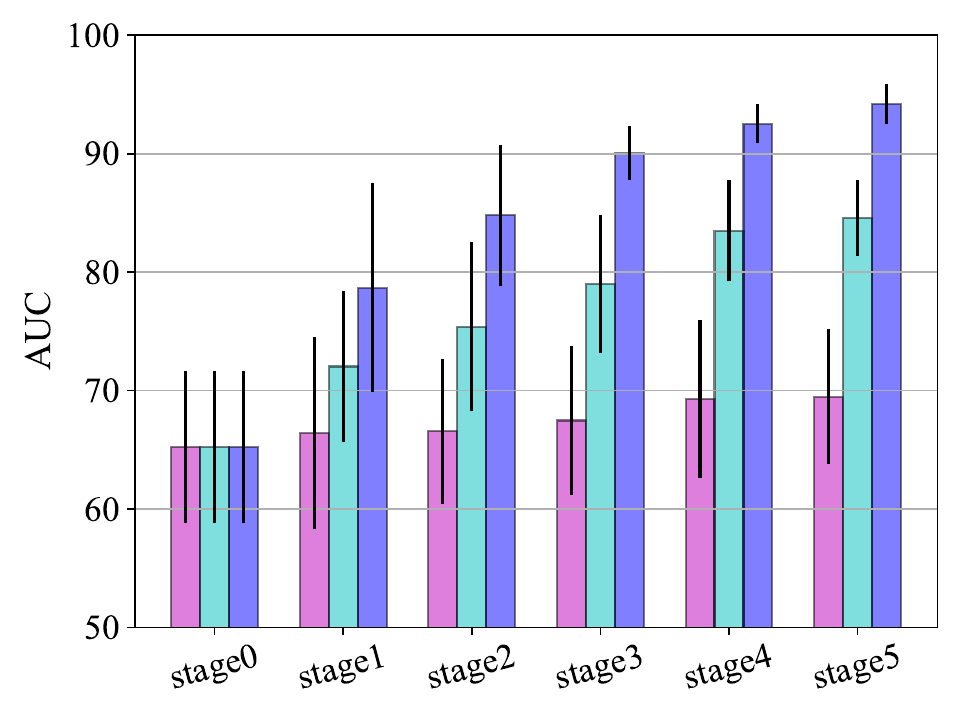}}
\subfigure[QS: Adaptive boundary\label{fig:agg-oc-f}]{\includegraphics[width=0.33\textwidth]{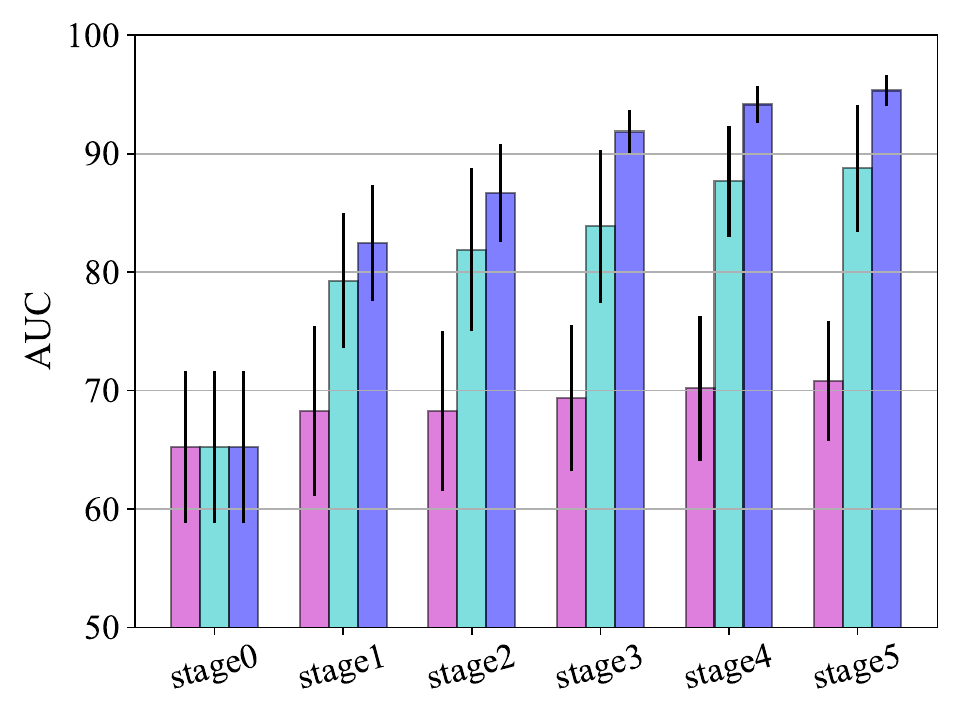}}
\caption{
    Aggregated results on all the datasets with anomaly detectors based on OC variants. 
    ((a), (b), (c)) Results according to QS with each SSL method. 
    ((d), (e), (f)) Results according to SSL method with each QS.
    We averaged the average AUC and a standard deviation computed over 5 seeds per dataset for the aggregation. 
    `stage0' means the initial stage.}
\label{fig:agg-oc}
\vspace{-3mm}
\end{figure*}

\vspace{2mm}\noindent\textbf{Competing methods.}\quad
We apply our methods to two versions of Deep SVDD (DSVDD)~\cite{ruff2018deep}, which are \textit{soft-boundary} DSVDD (SB) and \textit{One-Class} DSVDD (OC). 
Our query strategy (QS), which is Adaptive Boundary (AB), is compared with High-confidence (HC)~\cite{barnabe2015active, lesouple2021incorporating}, Decision-Boundary (DB)~\cite{gornitz2013toward}, and Random (RD). 
Note that DB can be applied only to SB because OC does not predict a decision boundary.
Our noise contrastive estimation (NCE)-based semi-supervised learning (SSL) method is compared with SB, OC, and DSAD~\cite{ruff2020deep}. 
We consider SB and OC as the SSL method that excludes labeled abnormal samples, and DSAD is the state-of-the-art SSL method based on OC.
We also compare our methods to a recent active learning approach for anomaly detection, named Unsupervised to Active Inference (UAI)~\cite{pimentel2020deep}, that transforms an unsupervised deep AD model into an active learning model by adding an anomaly classifier.

\vspace{2mm}\noindent\textbf{Implementation details.}\quad
To implement the base model, DSVDD\footnote{\url{https://github.com/lukasruff/Deep-SVDD-PyTorch}}, we use the source code released by the authors and adjust the backbone architectures for each dataset.
A 3-layer MLP with 32-16-8 units is used on the \textit{Ionosphere}, \textit{Cardio}, \textit{Glass}, \textit{Optdigits}, and \textit{NSL-KDD} dataset;
a 3-layer MLP with 128-64-32 units is used on the \textit{Arrhythmia} dataset; 
a 3-layer MLP with 64-32-16 units is used on the \textit{Mnist} dataset.
The model is pre-trained with a reconstruction loss from an autoencoder for 100 epochs and then fine-tuned with an anomaly detection loss for 50 epochs. 
The decoder of an autoencoder is implemented in a symmetric structure of the 3-layer MLP encoder.
We use Adam optimizer~\cite{kingma2015adam} with a batch size of 128 with a learning rate of $10^{-3}$.
Data samples are standardized to have zero mean and unit variance.
We set $B$ as 1\% of each dataset and $T$ as five.
In the case of small datasets where $n$ is less than 500 samples, \emph{e.g.}, ionosphere, arrhythmia, glass, 6 data samples are queried instead of 1\%.
The initial query boundary $q_1$ is set to 0.8.
The hyperparameter $\nu$ of SB is fixed to 0.5 for all the experiments.
For the hyperparameter $\eta$ of DSAD, we use 1, as suggested by the work~\cite{ruff2020deep}.
Following the structure of UAI~\cite{pimentel2020deep}, the UAI layer is attached to the last layer of an encoder.
For a fair comparison, we apply the same training scheme to UAI by pre-training the encoder with the autoencoder loss and then fine-tuning the encoder.
The experiments are performed with Intel Xeon Silver 4210 CPU and GeForce GTX 1080Ti GPU.

%% ---------------------------------------------------------------------------------------------------------------------------------------
\subsection{Experiment results}
\begin{figure*}[t]
    \centering
    \subfigure[SSL method: SB\label{fig:agg-sb-a}]{\includegraphics[width=0.35\textwidth]{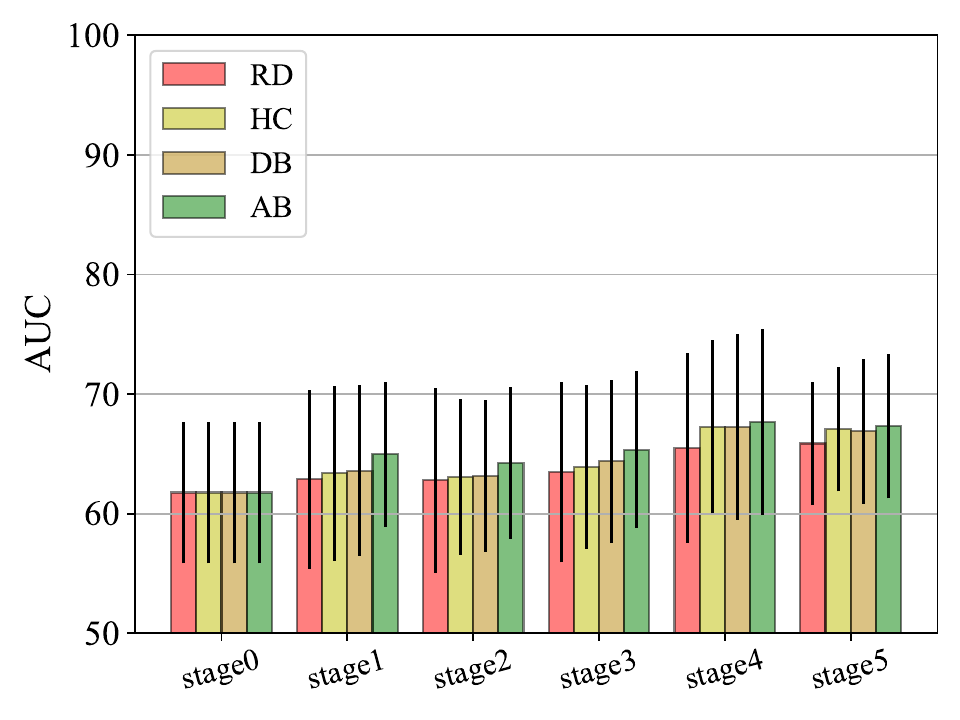}}
    \subfigure[SSL method: SB-NCE\label{fig:agg-sb-b}]{\includegraphics[width=0.35\textwidth]{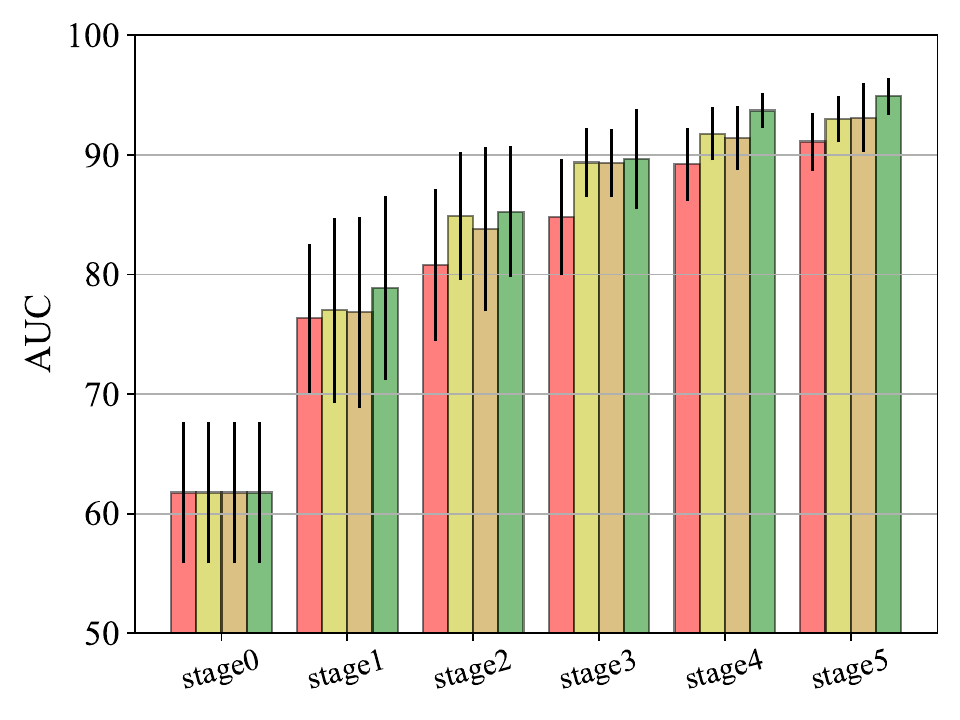}}
    \linebreak
    \subfigure[QS: Random\label{fig:agg-sb-c}]{\includegraphics[width=0.24\textwidth]{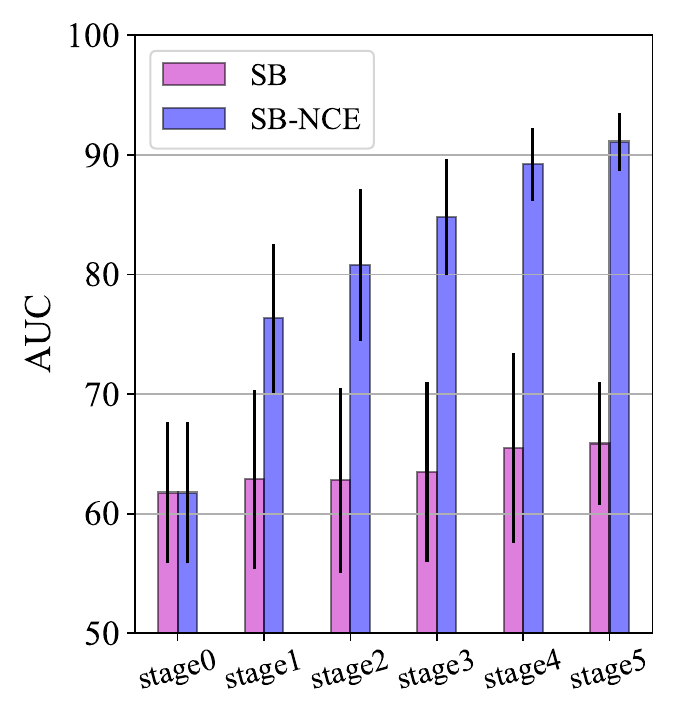}}
    \subfigure[QS: High confidence\label{fig:agg-sb-d}]{\includegraphics[width=0.24\textwidth]{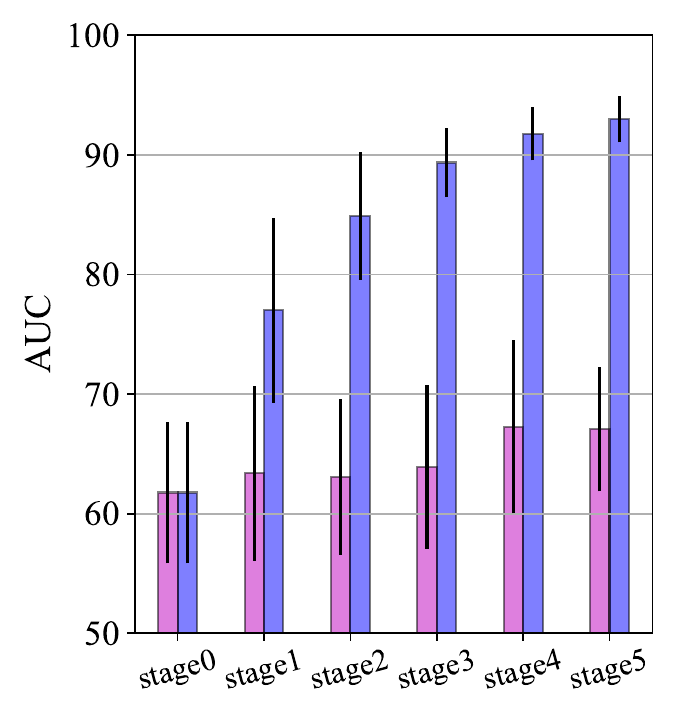}}
    \subfigure[QS: Decision boundary\label{fig:agg-sb-e}]{\includegraphics[width=0.24\textwidth]{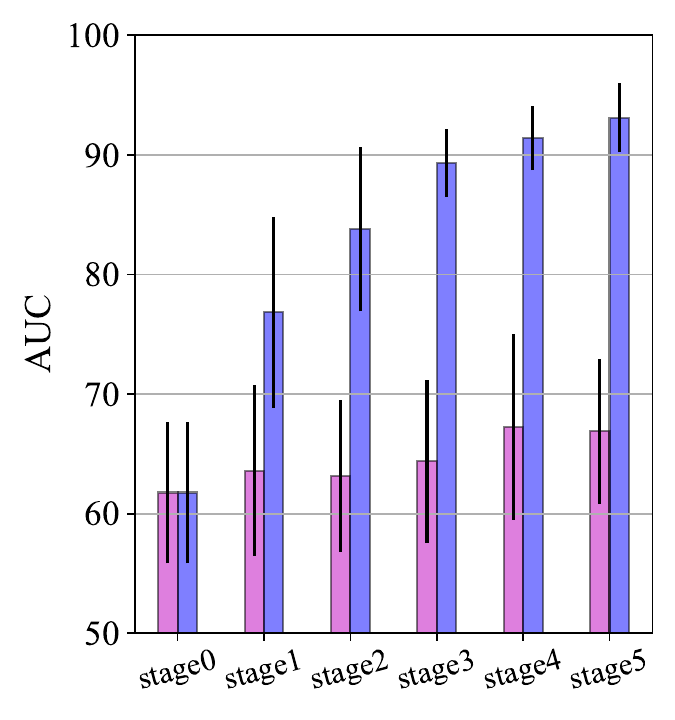}}
    \subfigure[QS: Adaptive boundary\label{fig:agg-sb-f}]{\includegraphics[width=0.24\textwidth]{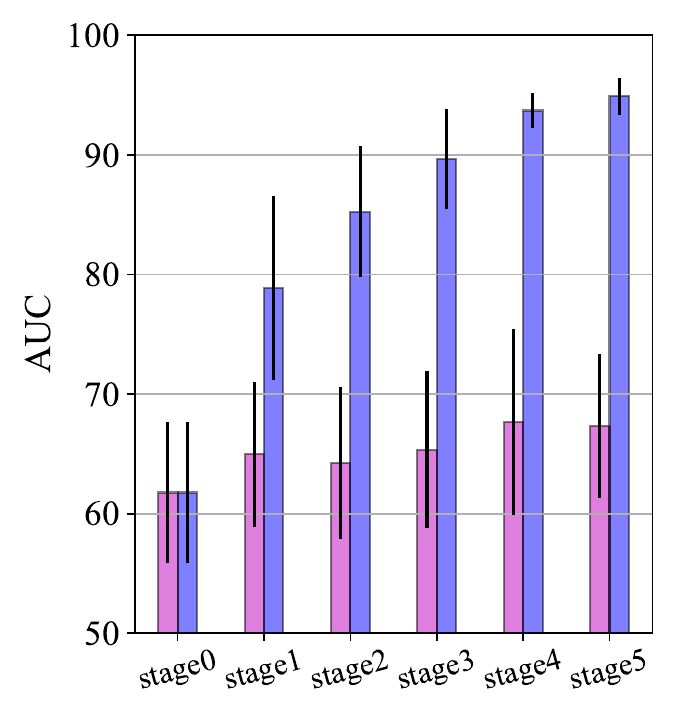}}
    \caption{
        Aggregated results on all the datasets with anomaly detectors based on SB variants. 
        ((a), (b)) Results according to QS with each SSL method. 
        ((c), (d), (e), (f)) Results according to SSL method with each QS.
        We averaged the average AUC and a standard deviation computed over 5 seeds per dataset for the aggregation. 
        `stage0' means the initial stage.}
    \label{fig:agg-sb}
\vspace*{-3mm}
\end{figure*}
\vspace{-2mm}
\noindent\textbf{Aggregated results.}\quad
To validate the effectiveness of the proposed QS and SSL method comprehensively, we show the aggregated results on all the datasets first (\Fref{fig:agg-oc} and \Fref{fig:agg-sb}). 
Each figure shows the results with OC (\Fref{fig:agg-oc}) and SB (\Fref{fig:agg-sb}) as a base model, respectively. 
AB shows the highest performance in most cases when used with SSL methods (\Fref{fig:agg-oc-a}, \Frefs{fig:agg-oc-c}, and \Fref{fig:agg-sb-a}, \Frefs{fig:agg-sb-b}). 
In particular, even in the case of excluding labeled abnormal samples during SSL, AB obtains higher performance than HC (\Fref{fig:agg-oc-a} and \Fref{fig:agg-sb-a}). 
It suggests that excluding highly uncertain abnormal samples by AB can be more conducive to normality learning than excluding highly confident abnormal samples by HC.
Similarly, AB shows higher performance than HC on SSL methods that use labeled abnormal samples for training (\Fref{fig:agg-oc-b}, \Frefs{fig:agg-oc-c}, and \Fref{fig:agg-sb-b}).
We observe that RD shows higher performance than AB when combined with DSAD (\Fref{fig:agg-oc-b}), while AB is more effective when combined with OC-NCE (\Fref{fig:agg-oc-c}).
It is due to the different use of labeled normal samples between DSAD and OC-NCE.
AB searches for a set of uncertain samples, including both normal and abnormal ones.
However, DSAD only repels labeled abnormal samples from the hypersphere center and ignores labeled normal samples. 
On the other hand, OC-NCE uses both labeled normal and abnormal samples by contrasting them, making the samples searched by AB more effective.
Furthermore, in the case of SB variants, AB shows higher performance than DB. 
Since DB relies on hyper-parameters and the best value may change for every active learning stage, it may fail to select informative samples.
In addition, the proposed NCE-based SSL method obtains the highest performance in most cases when used with any QS (\Fref{fig:agg-oc-d}, \Frefs{fig:agg-oc-e}, \Frefs{fig:agg-oc-f}, and \Fref{fig:agg-sb-c}, \Frefs{fig:agg-sb-d}, \Frefs{fig:agg-sb-e}, \Frefs{fig:agg-sb-f}), especially when combined with AB.

\begin{figure*}[t!]%[b]
    \subfigure[NSL-KDD (46.5\%)\label{fig:individual-oc-a}]{\includegraphics[width=0.245\textwidth]{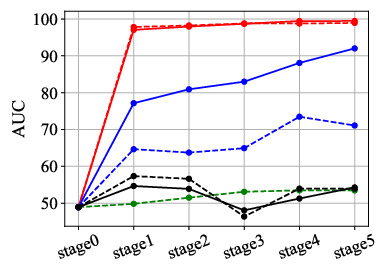}}
    \subfigure[Ionosphere (35.9\%)\label{fig:individual-oc-d}]{\includegraphics[width=0.245\textwidth]{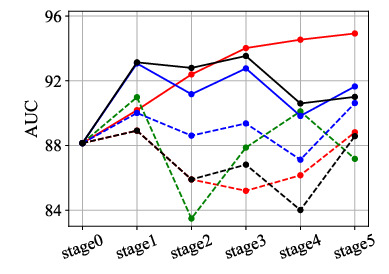}}
    \subfigure[Arrhythmia (14.6\%)]{\includegraphics[width=0.245\textwidth]{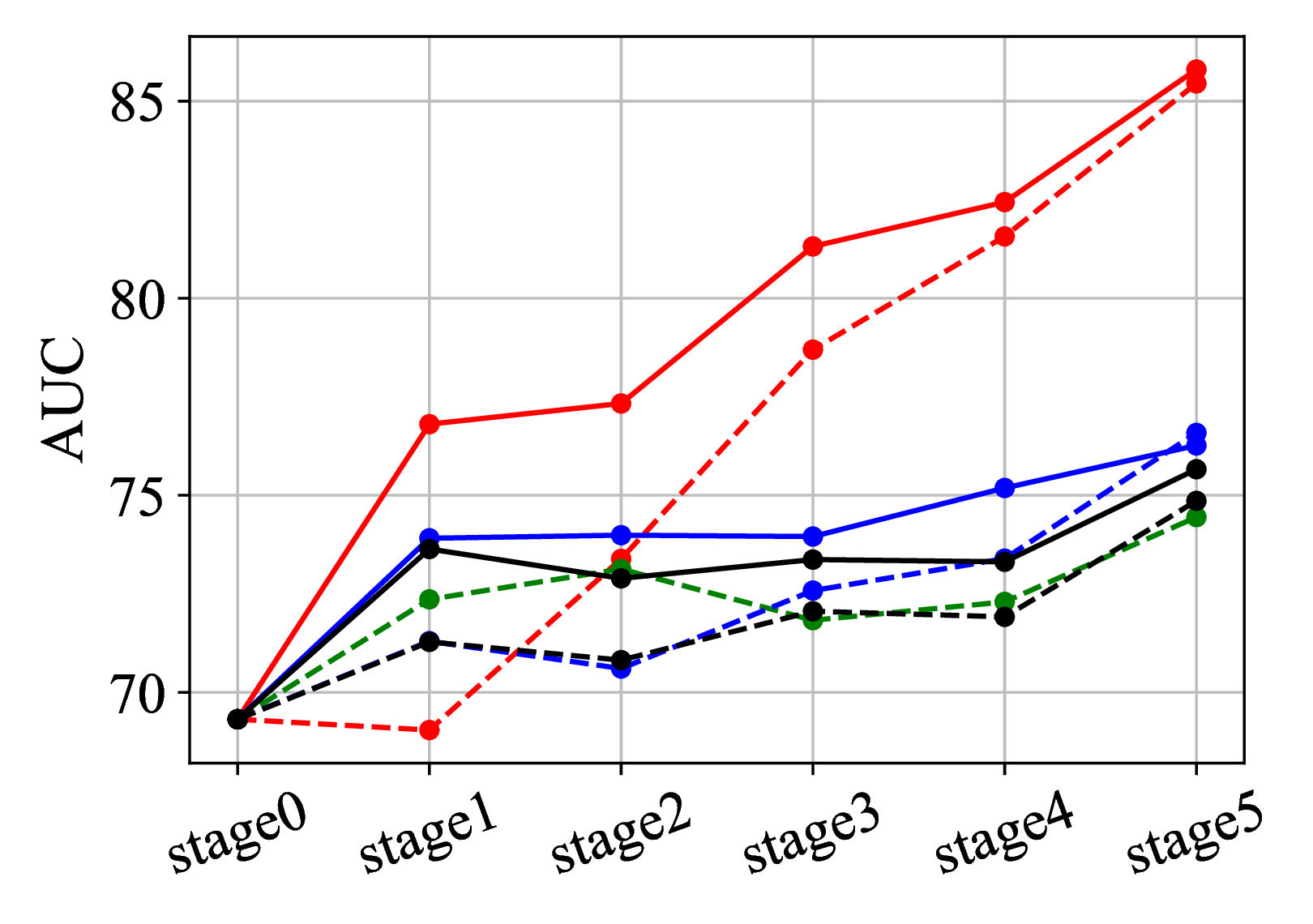}}
    \subfigure[Cardio (9.6\%)]{\includegraphics[width=0.245\textwidth]{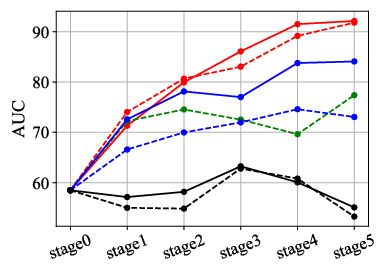}}
    \linebreak
    \subfigure[Mnist (9.2\%)]{\includegraphics[width=0.245\textwidth]{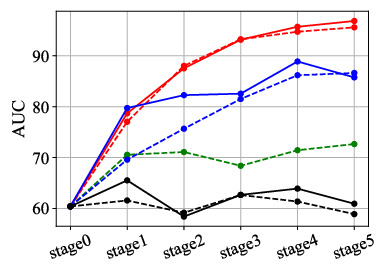}}
    \subfigure[Glass (4.2\%)]{\includegraphics[width=0.245\textwidth]{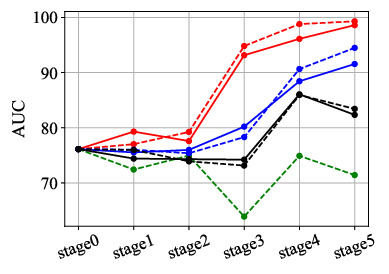}}    
    \subfigure[Optdigits (2.9\%)]{\includegraphics[width=0.245\textwidth]{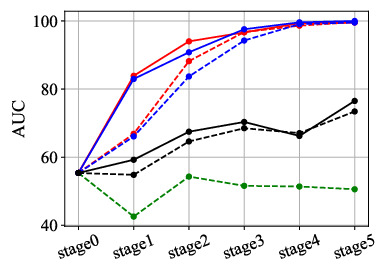}}
    \subfigure[\empty]{\includegraphics[width=0.245\textwidth]{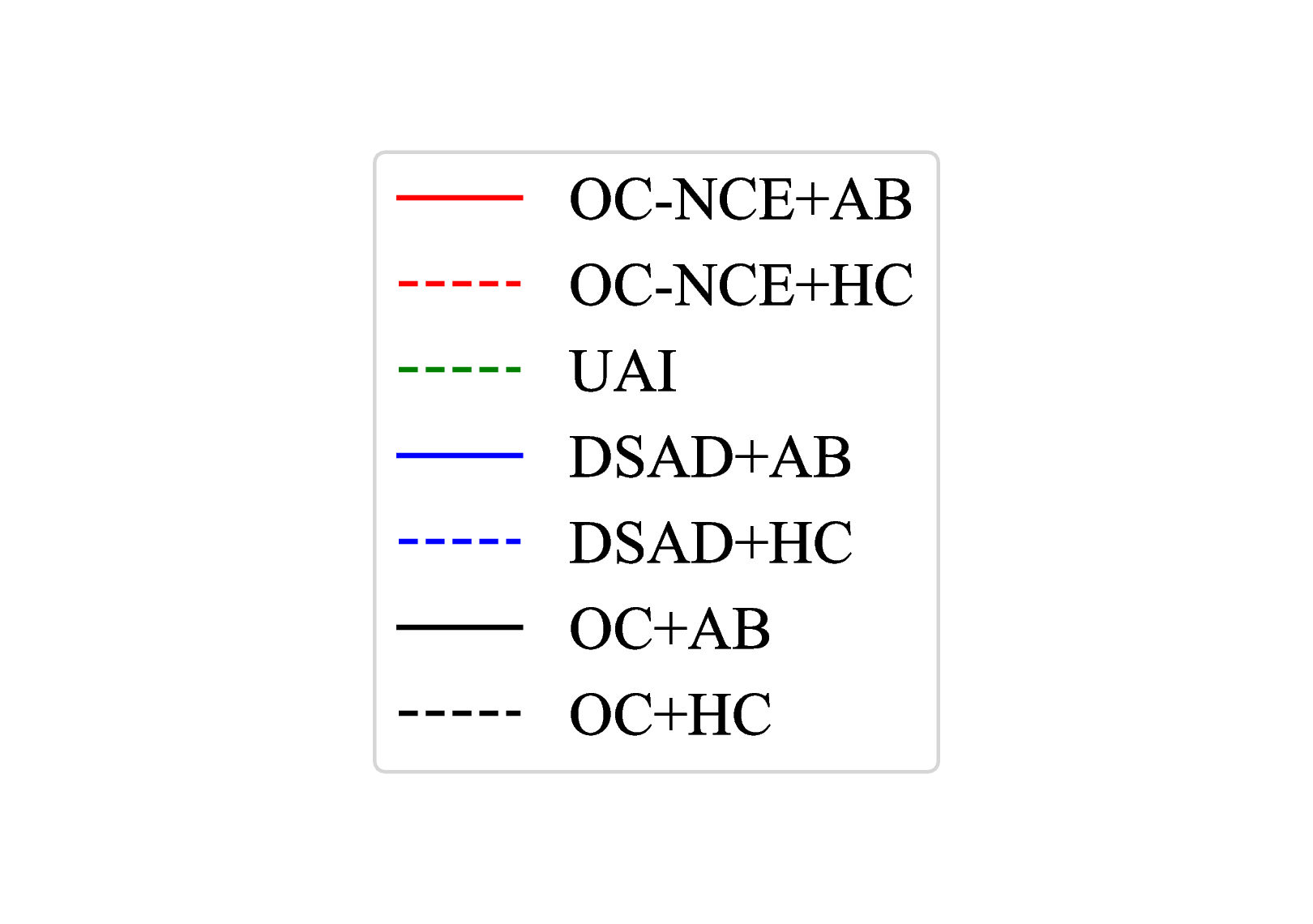}}
    \caption{
    The anomaly detection performance of OC variants according to active learning stages for each dataset.
    We report the average AUC computed over 5 seeds. `stage0' means the initial stage, and the numbers in parentheses indicate anomaly ratios.}
    \label{fig:individual-oc}
\end{figure*}

\vspace{2mm}\noindent\textbf{Results by dataset.}\quad
The anomaly detection performance with OC as a base model is shown for each dataset (\Fref{fig:individual-oc}). 
Random QS, the worst baseline among the competitors, is excluded for better visualization.
First, the same trend as in the aggregated results is observed: AB (solid lines) achieves higher performance in most cases for each SSL method, and the NCE-based SSL method (red lines) obtains higher performance than other SSL methods.
In addition, the difference among SSL methods is apparent in \Fref{fig:individual-oc}.
OC, which excludes labeled abnormal samples, has a smaller performance improvement or its performance fluctuates even as the number of labeled samples increases along with stages.
It indicates that the appropriate use of labeled abnormal samples is more effective for normality learning. 
Furthermore, compared to DSAD, considering only labeled abnormal samples, OC-NCE effectively improves performance by contrasting labeled normal and abnormal samples.
Meanwhile, UAI does not show stable performance improvement despite using both labeled normal and abnormal samples. 
UAI queries samples with high anomaly scores computed from the UAI layer.
Similar to HC, queried samples by UAI are biased to be abnormal samples and less effective than the proposed method.

Lastly, we compare the combination of OC-NCE and the two QSs, AB (solid red lines) and HC (dashed red lines). 
AB shows higher performance than HC favorably in the early stages in most cases. 
However, the performance with both QSs becomes similar while repeating active learning stages. 
It can be explained by AB gradually moving outward of the hypersphere's center as the model learns robust normality over the stages. 

\begin{figure*}[t!]
    \centering
    \subfigure[Changes in $r_t$ of OC-NCE+AB\label{fig:anal-rt-a}]{\includegraphics[width=0.95\textwidth]{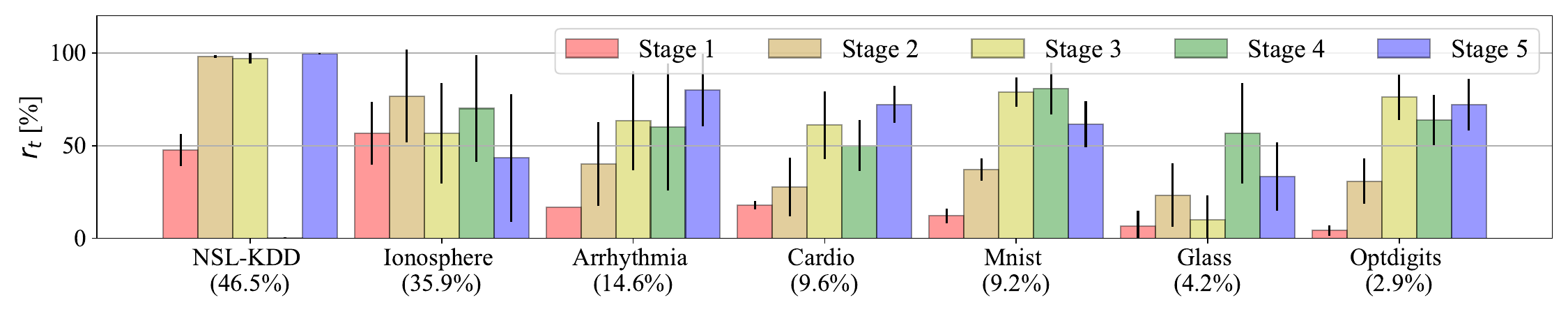}}
    \linebreak
    \subfigure[Changes in $q_t$ of OC-NCE+AB\label{fig:anal-rt-b}]{\includegraphics[width=0.95\textwidth]{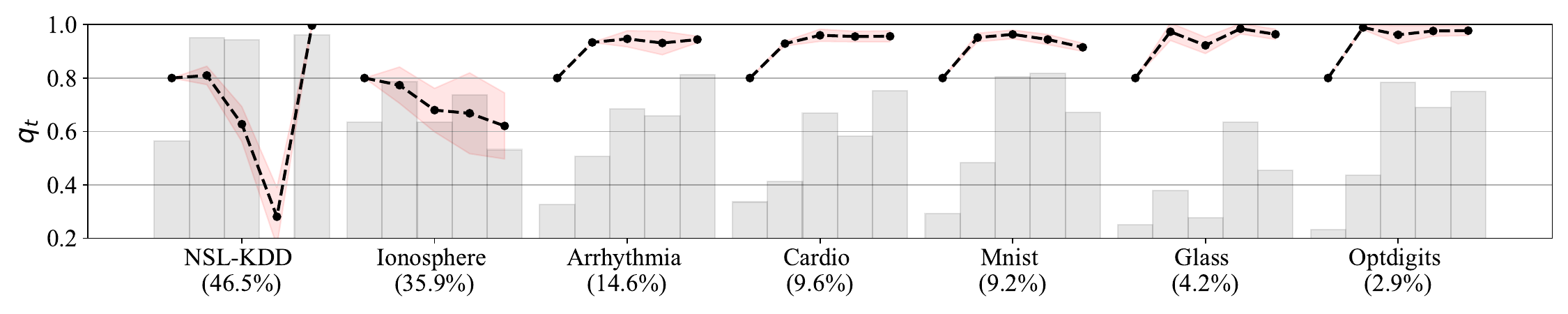}}
	\caption{
    Changes in (a) $r_t$ and (b) $q_t$ at each active learning stage of OC-NCE+AB on each dataset.
	We report the average value and standard deviation computed over 5 seeds.
	The large standard deviation on the \textit{Ionosphere}, \textit{Arrhythmia}, and \textit{Glass} dataset is presumed to be due to the small number of samples. The numbers in parentheses indicate anomaly ratios.}
	\label{fig:anal-rt}
	\vspace*{-3mm}
\end{figure*}

\vspace{2mm}\noindent\textbf{Analysis on queried samples.}\quad
We analyze $r_t$ and $q_t$ being updated at each active learning stage by OC-NCE+AB (\Fref{fig:anal-rt}) to show how uncertainty sampling works without learning a parameter-dependent boundary.
$r_1$ obtained from the trained model at the initial stage tends to be proportional to the anomaly ratio of each dataset (\Fref{fig:anal-rt-a}).
This result empirically shows that the model learns contaminated normality due to abnormal samples.
While repeating the active learning stages, the $r_t$ values go toward 50\%.
The latter stages tend to show a high abnormal sample ratio $r_t$. It is due to the improved model performance, resulting in more abnormal samples located near high $q_t$ regions.
Even if $r_t$ does not close to 50\% as expected, $q_t$ adaptively moves in the direction where $r_t$ is expected to be close to 50\%.
In a large anomaly ratio case, \eg, NSL-KDD, $q_t$ changes with large steps because $r_t$ values are far from 50\%. Although each $r_t$ is not close to 50\%, the accumulated labeled set becomes balanced as the active learning progresses.

%% ---------------------------------------------------------------------------------------------------------------------------------------
\section{Conclusion}\label{sec:Conclusion}
In this work, we present a method for active anomaly detection based on deep one-class classification.
The proposed query strategy selects samples closest to the boundary adjusted by the ratio of queried abnormal samples.
Unlike the previous uncertainty sampling based on a decision boundary, our query strategy does not require learning explicit boundary, which is controlled by hyper-parameters in one-class classification.
In addition, joint learning with noise contrastive estimation (NCE) is proposed as a semi-supervised learning method to effectively utilize labeled normal and abnormal data.
We address that the proposed query strategy and joint learning with NCE mutually enhance each other resulting in a favorable performance against the competing methods, including the state-of-the-art semi-supervised methods.

%% ---------------------------------------------------------------------------------------------------------------------------------------
\section*{Acknowledgments}
{\small
\noindent This work was supported by Institute of Information \& communications Technology Planning \& Evaluation (IITP) grant funded by the Korea government (MSIT) (No.2020-0-00833, A study of 5G based Intelligent IoT Trust Enabler).
The part of this work was done when Junsik Kim was in KAIST and supported by Basic Science Research Program through the National Research Foundation of Korea (NRF) funded by the Ministry of Education {\small (2020R1A6A3A01100087)}}.

\balance
\bibliographystyle{plain}  % Bibliography not compatible with author-year citations. 
\bibliography{reference}

\begin{thebibliography}{10}

\bibitem{nsl-kdd2018}
Nsl-kdd.
\newblock {\em URL https://www.unb.ca/cic/datasets/nsl.html}, 2018.

\bibitem{barnabe2015active}
Vincent Barnab{\'e}-Lortie, Colin Bellinger, and Nathalie Japkowicz.
\newblock Active learning for one-class classification.
\newblock In {\em IEEE International Conference on Machine Learning and Applications}, 2015.

\bibitem{chalapathy2018anomaly}
Raghavendra Chalapathy, Aditya~Krishna Menon, and Sanjay Chawla.
\newblock Anomaly detection using one-class neural networks.
\newblock {\em arXiv preprint arXiv:1802.06360}, 2018.

\bibitem{das2016incorporating}
Shubhomoy Das, Weng-Keen Wong, Thomas Dietterich, Alan Fern, and Andrew Emmott.
\newblock Incorporating expert feedback into active anomaly discovery.
\newblock In {\em IEEE International Conference on Data Mining}, 2016.

\bibitem{das2017incorporating}
Shubhomoy Das, Weng-Keen Wong, Alan Fern, Thomas~G Dietterich, and Md~Amran Siddiqui.
\newblock Incorporating feedback into tree-based anomaly detection.
\newblock {\em arXiv preprint arXiv:1708.09441}, 2017.

\bibitem{ghasemi2011active}
Alireza Ghasemi, Hamid~R Rabiee, Mohsen Fadaee, Mohammad~T Manzuri, and Mohammad~H Rohban.
\newblock Active learning from positive and unlabeled data.
\newblock In {\em IEEE International Conference on Data Mining Workshops}, 2011.

\bibitem{gornitz2013toward}
Nico G{\"o}rnitz, Marius Kloft, Konrad Rieck, and Ulf Brefeld.
\newblock Toward supervised anomaly detection.
\newblock {\em Journal of Artificial Intelligence Research}, 46:235--262, 2013.

\bibitem{gutmann2010noise}
Michael Gutmann and Aapo Hyv{\"a}rinen.
\newblock Noise-contrastive estimation: A new estimation principle for unnormalized statistical models.
\newblock In {\em International Conference on Artificial Intelligence and Statistics}, 2010.

\bibitem{kingma2015adam}
Diederik~P Kingma and Jimmy Ba.
\newblock Adam: A method for stochastic optimization.
\newblock In {\em International Conference on Learning Representations}, 2015.

\bibitem{lamba2019learning}
Hemank Lamba and Leman Akoglu.
\newblock Learning on-the-job to re-rank anomalies from top-1 feedback.
\newblock In {\em SIAM International Conference on Data Mining}, 2019.

\bibitem{lesouple2021incorporating}
Julien Lesouple and Jean-Yves Tourneret.
\newblock Incorporating user feedback into one-class support vector machines for anomaly detection.
\newblock In {\em IEEE European Signal Processing Conference}, 2021.

\bibitem{pang2021deep}
Guansong Pang, Chunhua Shen, Longbing Cao, and Anton Van~Den Hengel.
\newblock Deep learning for anomaly detection: A review.
\newblock {\em ACM Computing Surveys}, 54(2):1--38, 2021.

\bibitem{perera2021one}
Pramuditha Perera, Poojan Oza, and Vishal~M Patel.
\newblock One-class classification: A survey.
\newblock {\em arXiv preprint arXiv:2101.03064}, 2021.

\bibitem{pimentel2020deep}
Tiago Pimentel, Marianne Monteiro, Adriano Veloso, and Nivio Ziviani.
\newblock Deep active learning for anomaly detection.
\newblock In {\em 2020 International Joint Conference on Neural Networks (IJCNN)}, pages 1--8. IEEE, 2020.

\bibitem{rayana2016odds}
Shebuti Rayana.
\newblock Odds library.
\newblock {\em URL http://odds. cs. stonybrook. edu}, 2016.

\bibitem{ruff2021unifying}
Lukas Ruff, Jacob~R. Kauffmann, Robert~A. Vandermeulen, Gr{\'e}goire Montavon, Wojciech Samek, Marius Kloft, Thomas~G. Dietterich, and Klaus-Robert M{\"u}ller.
\newblock A unifying review of deep and shallow anomaly detection.
\newblock {\em Proceedings of the IEEE}, 109(5):756--795, 2021.

\bibitem{ruff2018deep}
Lukas Ruff, Robert Vandermeulen, Nico Goernitz, Lucas Deecke, Shoaib~Ahmed Siddiqui, Alexander Binder, Emmanuel M{\"u}ller, and Marius Kloft.
\newblock Deep one-class classification.
\newblock In {\em International Conference on Machine Learning}, 2018.

\bibitem{ruff2020deep}
Lukas Ruff, Robert~A. Vandermeulen, Nico G{\"o}rnitz, Alexander Binder, Emmanuel M{\"u}ller, Klaus-Robert M{\"u}ller, and Marius Kloft.
\newblock Deep semi-supervised anomaly detection.
\newblock In {\em International Conference on Learning Representations}, 2020.

\bibitem{sabokrou2018adversarially}
Mohammad Sabokrou, Mohammad Khalooei, Mahmood Fathy, and Ehsan Adeli.
\newblock Adversarially learned one-class classifier for novelty detection.
\newblock In {\em IEEE Conference on Computer Vision and Pattern Recognition}, 2018.

\bibitem{scholkopf2001estimating}
Bernhard Sch{\"o}lkopf, John~C Platt, John Shawe-Taylor, Alex~J Smola, and Robert~C Williamson.
\newblock Estimating the support of a high-dimensional distribution.
\newblock {\em Neural computation}, 13(7):1443--1471, 2001.

\bibitem{tax2004support}
David~MJ Tax and Robert~PW Duin.
\newblock Support vector data description.
\newblock {\em Machine learning}, 54(1):45--66, 2004.

\bibitem{trittenbach2020overview}
Holger Trittenbach, Adrian Englhardt, and Klemens B{\"o}hm.
\newblock An overview and a benchmark of active learning for outlier detection with one-class classifiers.
\newblock {\em Expert Systems with Applications}, 168:114372, 2020.

\bibitem{van2008visualizing}
Laurens Van~der Maaten and Geoffrey Hinton.
\newblock Visualizing data using t-sne.
\newblock {\em Journal of Machine Learning Research}, 9(86):2579--2605, 2008.

\bibitem{xia2015learning}
Yan Xia, Xudong Cao, Fang Wen, Gang Hua, and Jian Sun.
\newblock Learning discriminative reconstructions for unsupervised outlier removal.
\newblock In {\em IEEE International Conference on Computer Vision}, 2015.

\bibitem{zha2020meta}
Daochen Zha, Kwei-Herng Lai, Mingyang Wan, and Xia Hu.
\newblock Meta-aad: Active anomaly detection with deep reinforcement learning.
\newblock In {\em IEEE International Conference on Data Mining}, 2020.

\bibitem{zhang2021anomaly}
Zheng Zhang and Xiaogang Deng.
\newblock Anomaly detection using improved deep svdd model with data structure preservation.
\newblock {\em Pattern Recognition Letters}, 148:1--6, 2021.

\end{thebibliography}

\end{document}